\documentclass{article}
\usepackage{ijcai20}

\usepackage{times}

\usepackage{soul}
\usepackage{url}
\usepackage[hidelinks]{hyperref}
\usepackage[utf8]{inputenc}
\usepackage[small]{caption}
\usepackage{graphicx}
\usepackage{amsmath}
\usepackage{amsfonts}
\usepackage{booktabs}
\title{Tensor Network for Supervised Learning at Finite Temperature}
 \author{
 Haoxiang Lin$^{1}$\and
Shuqian Ye$^{1}$\and
Xi Zhu$^{1}$\footnote{Contact Author}\\
\affiliations
$^1$Shenzhen Institute of Artificial Intelligence and Robotics for Society, \\
the Chinese University of Hong Kong, Shenzhen\\
\emails
\{haoxianglin, shuqianye\}@link.cuhk.edu.cn,
xizhu@cuhk.edu.cn
}

\begin{document}

\maketitle

\begin{abstract}
The large variation of datasets is a huge barrier for image classification tasks. In this paper, we embraced this observation and introduce the finite temperature tensor network (FTTN), which imports the thermal perturbation into the matrix product states framework by placing all images in an environment with constant temperature, in analog to energy-based learning. Tensor network is chosen since it is the best platform to introduce thermal fluctuation. Different from traditional network structure which directly takes the summation of individual losses as its loss function, FTTN regards it as thermal average loss computed from the entanglement with the environment. The temperature-like parameter can be automatically optimized, which gives each database an individual temperature. FTTN obtains improvement in both test accuracy and convergence speed in several datasets. The non-zero temperature automatically separates similar features, avoiding the wrong classification in previous architecture. The thermal fluctuation may give a better improvement in other frameworks, and we may also implement the temperature of database to improve the training effect. 
\end{abstract}
\section{Introduction}
Originated from quantum many-body physics and quantum information sciences, tensor networks (TNs) states are building blocks designed for the manipulation of very high-dimensional data, representing the complex quantum states \cite{kuhn2019combined}. Key information of the overall wave function is encoded in individual block, and the insights into the entire framework enable the powerful numerical simulation approaches \cite{orus2019tensor,orus2014practical}. In the 2000s, TNs are applied in quantum information theory to explore many-body entanglement in the low-energy eigenstates, like the ground state quantum spin chain in 1D \cite{fannes1992finitely}, a lattice model on a thin 2D torus \cite{milsted2019tensornetwork}, and also excited states like exciton and bi-exciton \cite{kuhn2019combined}. Anywhere there is a correlation, there is also room to apply TNs' ability of estimating many-body entanglement.

Now, TNs are active in interdisciplinary areas besides physics. In recent years, great progress has been made in applying tensor networks to machine learning for both supervised and unsupervised tasks: the matrix product state (MPS) can be used as classifier \cite{stoudenmire2016supervised}, the tree tensor network (TTN) is equivalent to a deep convolutional arithmetic circuit (ConvAC) \cite{levine2017deep}
, and both MPS and TNN can be designed as generative models \cite{han2018unsupervised,cheng2019tree}. Meanwhile, TN has been investigated for other machine learning applications like the compressing of neural network weight layers \cite{novikov2015tensorizing}, the compressed sensing \cite{ran2019quantum} and data completion \cite{wang2016tensor}. TN is proven to be one of the most suitable platforms in connecting physics and machine learning, which is the reason why we choose TN as the platform to insert thermal fluctuation.

One excellent optimization method for MPS classifier is adapted from density matrix renormalization group (DMRG) algorithm in physics. \cite{stoudenmire2016supervised} Besides this DMRG-like optimization algorithm, automatic gradient can also be implied to network training \cite{efthymiou2019tensornetwork}. However, despite great achievements, they use a tensor network structure corresponding to physical system at high temperature, all states are calculated. 
If the variation between graphs from the same label is small and that from different labels is large, all features have large separation, so the analysis of all states is proper and the noise does not matter a lot. However, when the variation between images from the same label is relatively large and that from different labels is narrowed, the inputs are becoming complex so that we need to focus on the main feature. The proposed solution is thermal fluctuation,  which requires us to find the typical wave function at finite temperature. In physics, one useful approach is the Minimally Entangled Typical Quantum States (METTS), which ensemble the average of pure states to obtain an excellent approach. A finite-T DMRG algorithm is also provided to obtain the states at finite temperature \cite{stoudenmire2010minimally}. 

In this work, we combine the finite temperature systems with tensor network, by introducing an extra temperature layer. In Section~\ref{sec.TNandTL}, we construct the tensor network structure at finite temperature (FTTN), and its corresponding optimization algorithm is described in Section~\ref{sec.OPT}. Then we give a physical interpretation in Section~\ref{sec.PhyInter}. To summarize the result as depicted in Section~\ref{sec.result}, we find that, by  introducing a temperature layer into the MPS, similar features can be separated, giving improvement on the test accuracy and convergence speed. For large variation dataset, Fashion-MNIST  \cite{Fashion-MNIST}, the test accuracy is increased from $87.73\%$ to $88.72\%$, around one percentage increment. Even for the small variation dataset, MNIST \cite{mnist}, the test accuracy is increased from 98.31\% to 98.43\%. This method can gives rise to other machine learning frameworks like graph neural network (GNN) and convolutional neural network (CNN).
\section{Tensor Network and Temperature Layer}
\label{sec.TNandTL}
In this section, we briefly introduce the relationship between MPS for quantum physics calculation and MPS for machine learning. Then we propose our temperature-endorsed tensor network structure for machine learning.  We will concretely interpret our FTTN framework later in Section~\ref {sec.PhyInter}.

\begin{figure}[htb] 
    \center{\includegraphics[width=8cm]  {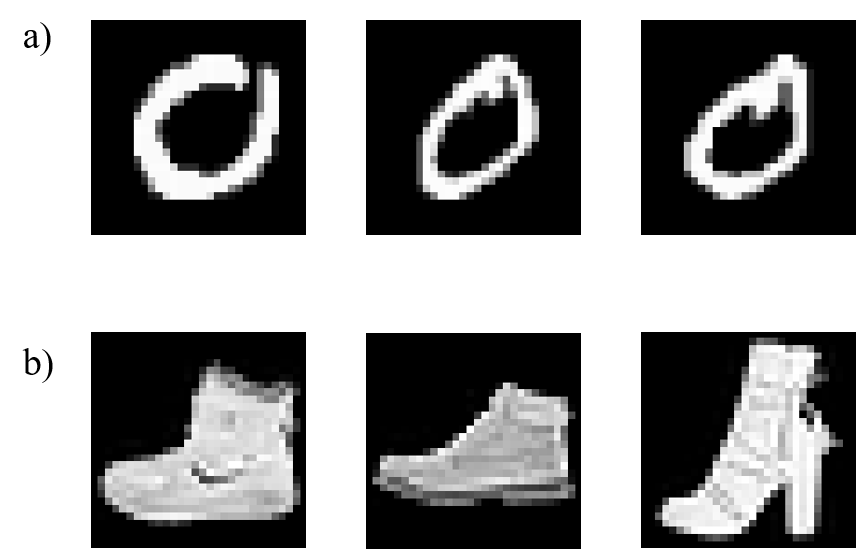}} 
    \caption{\label{fig.Illustration} Images with same label from different datasets. a) images with label ``0''  from MNIST dataset. b) images with label ``ankle boot'' 
     from fashion-MNIST dataset. It can be seen that the first three images are very similar, but the latter three have large variation.} 
\end{figure} 

In this work, we focus on the image classifications task. One dataset that traditional neural network structure performs badly is the large variation dataset. As shown in Figure~\ref{fig.Illustration}, the first three images are adapted from MNIST with the same label ``0''. They share the same shape, a feature ``circle'' is enough to describe it. The traditional networks give larger than 97\% \cite{stoudenmire2016supervised} without a complex framework. However, when it comes to large variation datasets like fashion-MNIST as shown in Figure~\ref{fig.Illustration}b, state of the art deep learning methods obtaining about 93\% test accuracy \cite{bhatnagar2017classification}. A huge difference exists between these images even though they come from the same label ``ankle boot''. It is hard to describe them simply using one feature. Compared with the first image, the second one are shaper and the third image has an extra heel. Even though human can notice these features, it is harder for network structure to distinguish their similarity.

\begin{figure}[htb] 
    \center{\includegraphics[width=8cm]{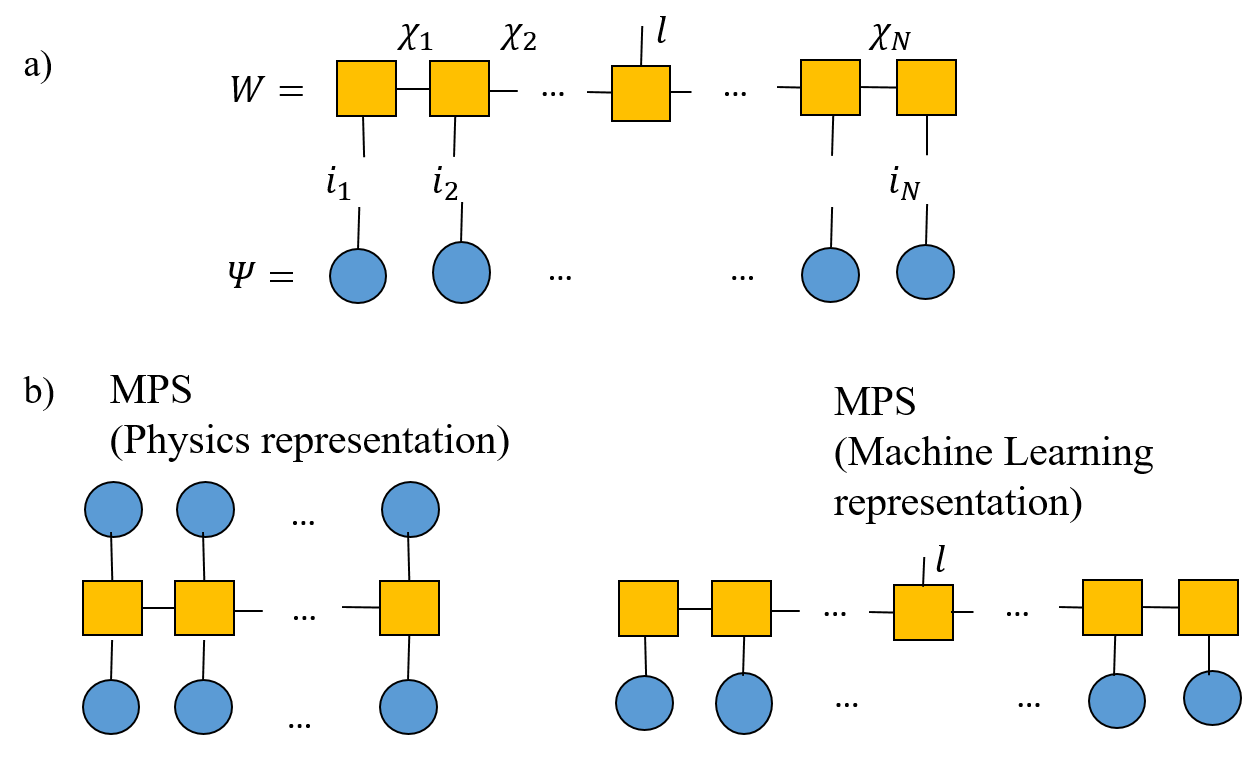}} 
    \caption{\label{fig.MPS} a) The tensor network representation of weight tensor $\mathsf W$ and feature map $\boldsymbol \Psi$. Each pixel of an image is encoded in a rank-1 tensor, also known as vector, corresponding to Equation \ref{eqn.feature_map}. The entire weight tensor $\mathsf W$ is approximated by contracted sequence of low-order tensors $\mathsf A$, corresponding to Equation \ref{eqn.weight}. b) MPS for quantum physics and machine learning. The left sub-figure stands for the expected value of an observable, while the right sub-figure is the tensor network architecture for machine learning. } 
\end{figure}

Though MPS is specialized in estimating one-dimensional correlation, it can also be applied to higher-dimensional data as well, where the two-dimension image is mapped into a one-dimensional chain.
The non-linear kernel learning requires the manipulation of very large tensor, where the input data $\boldsymbol X$ are mapped into a high dimension space by a feature map $\Psi$ before the final decision $f(\boldsymbol X) = \mathsf W \cdot \Psi(\boldsymbol X)$. To deal with the large tensors, the kernel trick is adapted, which only requires working with scalar products of feature tensors \cite{muller2001introduction}. To apply TN to machine learning, a feature map $\Psi$ in the form of local map $\psi$ multiplication
\begin{equation}
\boldsymbol \Psi(\boldsymbol X)=\boldsymbol \Psi^{(\boldsymbol S)}(\boldsymbol p)=\boldsymbol \psi^{(S_1)}(p_1) \otimes \boldsymbol \psi^{(S_2)}(p_2) \otimes \cdots \otimes \boldsymbol \psi^{(S_N)}(p_N)
\label{eqn.feature_map}
\end{equation}
is adapted, where $\boldsymbol S = [S_1,S_2,...,S_N]$ is the pixel position and $\boldsymbol p=[p_1,p_2,...,p_N]$ is its corresponding gray scale value ranging from 0 for black to 1 for white, $\otimes $ is the Kronecker product. One feature map used in \cite{stoudenmire2016supervised} is $\boldsymbol \psi(p)=[\cos(\pi p/2) \quad \sin(\pi p/2)]^T$, mapping the gray-scale value to a quantum spin. Another feature map is $\psi(x)=[1-p \quad p]^T$, which is used in \cite{efthymiou2019tensornetwork}. Sometimes this column vector is written as $\left | \boldsymbol \psi \right \rangle$ and its conjugate transpose is written as $\left \langle \boldsymbol \psi \right |$. But the selection of feature map is not important. These local map $\boldsymbol \psi$ correspond to individual blue circle with an edge as shown in Figure~\ref{fig.MPS}(a).  For a certain learning task and a specified feature map, a MPS $\mathsf W$ (contracted sequence of low-order tensors $\mathsf A^{(S)}$, the yellow cubic shown in Figure~\ref{fig.MPS}b) can approximate weight tensor , 
\begin{equation}
\mathsf W^{(\boldsymbol S)}=\sum_{\chi_1,\chi_2,...,\chi_N} \mathsf A_{i_1 \chi_1}^{(S_1)} \mathsf A_{i_2 \chi_1 \chi_2}^{(S_2)} \cdots \mathsf A_{i_N \chi_N}^{(S_N)}\label{eqn.weight}
\end{equation}
the subscript of $\mathsf A^{(S)}$ is its edges and  the connecting edge between nearest tensors is the bond dimensions $\chi$, one hyper-parameter of the tensor network. Figure~\ref{fig.MPS} indicates the structure. The lines represents tensor edges and the sharing edges indicates summation. For example, the contraction of left-most weight tensor $\mathsf A_{i_1 \chi_1}^{(S_1)}$ and feature map tensor $\boldsymbol \psi^{(S_1)}(p_1)$ gives $\sum_{i_1} \mathsf A_{i_1 \chi_1}^{(S_1)} \boldsymbol \psi^{(S_1)}(p_1)$, a rank 2 tensor, also known as matrix. The label index $l$ is placed arbitrarily at one tensor and this tensor is named as label tensor. If the weight matrix and the feature map are connected, the entire tensor network represents the final decision $f(\boldsymbol X) = \mathsf W \cdot \boldsymbol \Psi(\boldsymbol X)$.

The MPS framework in Figure~\ref{fig.MPS}(b) is in analog to calculating the average of an observable when the feature map only consists of real number. Originally in physics, the expectation of an observable is $\left \langle \boldsymbol \psi \right | \mathsf H \left |\boldsymbol \psi \right \rangle$, where the Hamiltonian $\mathsf H$ is a transpose conjugate matrix standing for the observable, similar to the weight layer.  Indices connected to the wave functions $\left \langle\boldsymbol \psi \right |$ and $\left | \boldsymbol \psi \right \rangle$ are generally the same, or sometimes conjugated. \cite{fannes1992finitely,klumper1993matrix,orus2014practical}. As for the machine learning task, one of these two indices is truncated as shown in Figure ~\ref{fig.MPS}(b). This is the same MPS structure as \cite{stoudenmire2016supervised,efthymiou2019tensornetwork}, if we treat the observable $\mathsf H$ as a modified weight tensor.

\begin{figure}[htb] 
    \center{\includegraphics[width=8cm]  {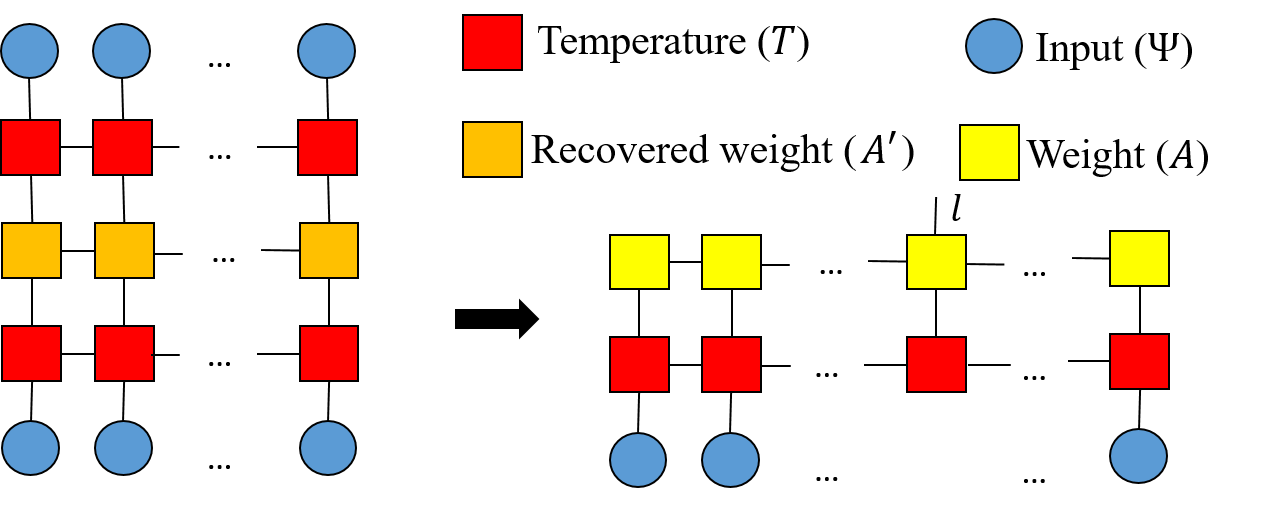}} 
    \caption{\label{fig.TNrep} The evolution of our tensor network for supervised learning at finite temperature. The left sub-figure is the physical representation of calculating observables with the endorsement of the temperature layer. The middle sub-figure is the simplified version, where we truncate one edge to reduce the number of tensors. The right sub-figure contract the weight tensor with its corresponding temperature tensor to obtain a simplified version, which is in analog to the previous MPS framework.} 
\end{figure}

Inspired by the very successful Minimally Entangled Typical Quantum States (METTS) algorithm \cite{stoudenmire2010minimally}, we propose an MPS framework at finite temperature. First, we need to recover the original tensor network structure with two input wave functions as shown in the left sub-figure of Figure~\ref{fig.MPS}b. The diagonal element of $\mathsf A'_{:,ij,:}$ is set as the element of $\mathsf A_{:,ij}$, where $i$ and $j$ are the inter-layer edges, this procedure adds an extra edge. Notice that now the weight tensor $\mathsf A^{(s)}$ has the rank of 4 except the boundary tensor which has the rank of 3. Then the temperature layer $\mathsf T$ is constructed by taking the exponential of recovered weight tensors $-\beta \mathsf A'$ where $\beta$ is a temperature-like constant, which is $\mathsf T = \exp(-\beta \mathsf A')$.
 The upper and lower edges of the temperature layer are connected to the input feature map and the weight layer. This recovery step is necessary since in machine learning representation, the rank of weight matrix is 3 except the boundary tensor which has the rank of 2. If we simply take exponential of these weight matrices, the number of edges is not enough to connect both the weight layer and the feature map. From this step, the tensor network structure is formulated as shown in the left sub-figure of Figure~\ref{fig.TNrep}. Then we use the initial weight tensor to replace this modified weight tensor to truncate the upper part of this framework, providing a simplified version as shown in the right sub-figure of Figure~\ref{fig.TNrep}. Now, FTTN is constructed.
 
In this work, we are interested in classifying data with given hidden labels. The contraction of the entire FTTN framework gives a vector, whose largest element corresponds to the output class.

\section{Optimization Algorithm}
\label{sec.OPT}
In this section, we focus on the optimization algorithm of FTTN. It is a slightly modified version based on gradient descent.

As for the Bayesian neural network, there exists another way to incorporate temperature effect. \cite{baldock2019bayesian} This method defines the potential energy of the network according to the calculated loss, and then the posterior distribution is temperature-adjusted based on input batches and the potential energy. 
However, in the FTTN, the potential energy is based on the weight tensor $\mathsf A$ instead of the output from the entire network.
Such a method can be useful, but in this work, we use another algorithm. Here the loss function is set as multi-class cross-entropy defined as:
\begin{equation*}
\text{Loss}=-\sum_{(\boldsymbol x_i,\boldsymbol y_i)} \log\left(  \text{Softmax}\left( f^{(\boldsymbol y_i)} (\boldsymbol x_i)\right)\right)
\label{eqn.LossFunc}
\end{equation*}

The gradient of each individual tensor $\mathsf A^{(S)}$ can be computed through a ``sweeping'' optimization algorithm \cite{stoudenmire2016supervised} or automatic gradients which is built-in to TensorFlow \cite{roberts2019tensornetwork,efthymiou2019tensornetwork}. By adding the coefficient to the obtained gradient, FTTN can be optimized easily without great change. After revising the weight layer, the temperature layer is simultaneously adjusted. Other optimization methods can also be applied to FTTN as long as its gradient can be computed and adjusted based on our proposed method.

Due to the endorsement of temperature layer, the back propagation process is modified. The exact solution is complex, and we propose an approximation. Notice that the temperature layer ($\mathsf T$) and the weight layer ($\mathsf W$) are strongly related by the formula $\mathsf T=\exp (-\beta \mathsf W )$. We first contracted the temperature layer with its corresponding weight layer. 
This contraction produces a T-dependent weight layer $\hat{\mathsf O}$:
\begin{equation}
\hat{\mathsf O} = \mathsf A\cdot \exp(-\beta\cdot \mathsf A)
\end{equation}\label{eqn.defTW}\\
Therefore, in the back-propagation process, an coefficient $c$ need to be added:
\begin{equation}
\mathsf C=\frac{\partial \hat{\mathsf O}}{\partial \mathsf A}=(1-\beta \mathsf A)\cdot \exp(-\beta \mathsf A)
\label{eqn.ApproGrad}
\end{equation}
Compared with the traditional back-propagation method for TN, the DMRG-like optimization method, the adjustment is the change of gradient. This is in analogue to modify the gradient matrix by multiplying the coefficient $c_{ijk}$ on each element of the matrix. For the other built-in optimizer of TensorFlow, we can also directly change the gradient matrix to optimize FTTN. 

%

\begin{figure}[htb] 
    \center{\includegraphics[width=8cm]{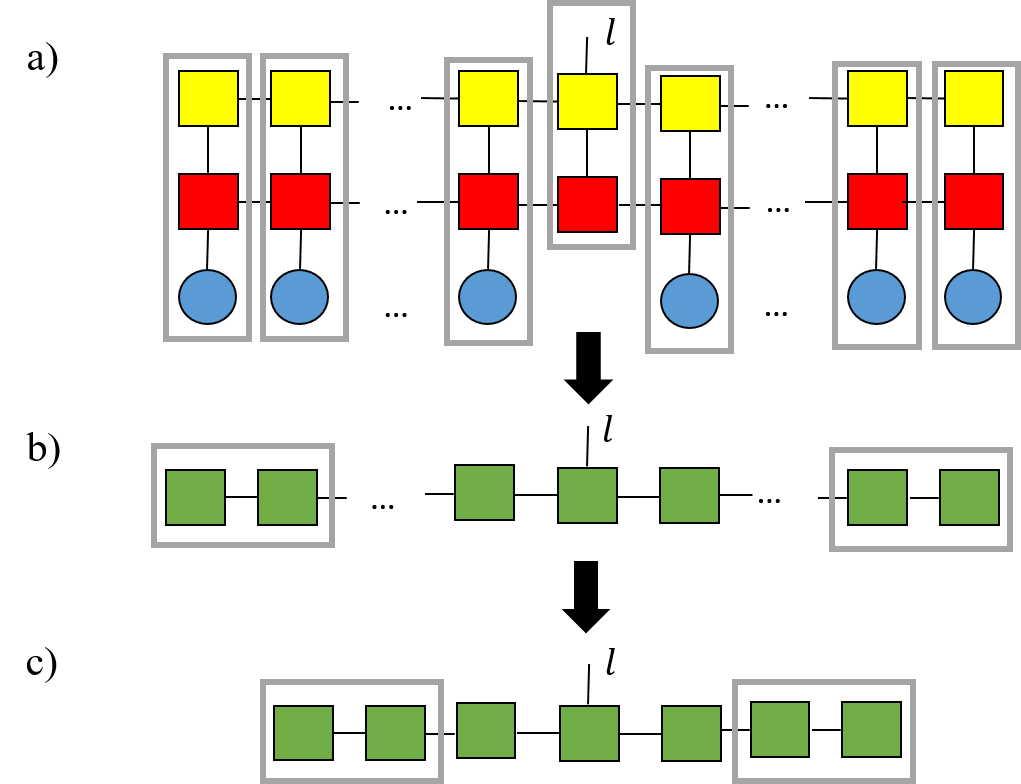}} 
    \caption{\label{fig.Imple} The three steps contraction order of FTTN. (a) Step 1, contract temperature tensor with its corresponding weight tensor and local feature map. This creates new effective tensors depicted as green squares. (b) Step 2, contract the generated tensors in pairs. This step is conducted independently for each pair of them in parallel. (c) Step 3, repeat step 2 until the tensor chain is fully contracted.}
\end{figure}

Even though the sweeping algorithms like the DMRG are preferred in physical applications due to its fast convergence, the simplicity of gradient-based optimization are more attractive to machine learning applications, which is the method we implied. In this FTTN framework, after calculating the gradient matrix, we calculate the coefficient matrix based on equation~\ref{eqn.ApproGrad}. Then the coefficient matrix is used to adjust the gradient. The temperature layer is adjusted based on the weight layer.

The contraction order matters the computation complexity. The DMRG-like order gives the total time complexity of order $O(NLd^2\chi^2)$\cite{stoudenmire2016supervised}, but it can not be forwarded to parallelized computation. Here we use a parallelized contraction order. The first step contracts the temperature layer with its corresponding weight layers and the feature map. This step gives a tensor chain. In the second step, we contract the nearest tensors in pair, then this step is repeated until the tensor chain is fully contracted. 
The total time complexity of this contraction is  $O(Nd^2\chi^2+\chi^3log(N))$. Even though the total cost is near the same as the previous one, this contraction order has the advantage that each step is parallelized as matrix multiplication without requiring the calculating result from the neighboring calculation. We noticed that the parallelized computation gives great enhancement on calculation speed.

\section{Physical Interpretation}
\label{sec.PhyInter}
In this section, we will briefly investigate the MPS framework and the effect of the endorsed temperature layer. This layer gives a thermal perturbation to the feature map, flatten the noise. The dataset with large variation contains large noise, and the endorsement of the temperature layer flatten the noise to obtain a better performance.

To summarize our tensor network framework, we embed data into space in the form of tensor, then the inner product between data and weight gives the final prediction. It should be noticed that the MPS tensor $\mathsf A$ can be interpreted as another vector in the tensor space, intuitively, a combination of all images in a given class. Ideally, for a specific class, an image not belonging to this class should be orthogonal to the MPS tensor, while an image belonging to this class has a large overlap with the MPS tensor. In physics, this is in analog to calculate the probability for any outcome of well-defined observables, which is calculated through the density matrix $\rho$. One may define the loss function as potential energy, and now the optimization is in analog to obtain the ground states. Different from traditional network structure which directly takes the summation of individual losses as its loss function, the input feature goes through a temperature layer to be entangled with environment, in the other words, FTTN regards the loss as thermal average loss computed from the entanglement with the environment.

The insertion of the temperature layer is inspired by the METTS algorithm in physics applications \cite{white2009minimally}. The fundamental proposition of statistical mechanics indicates that the density matrix of a system at inverse temperature $\beta$ with Hamiltonian $\mathsf H$ is $\mathsf \rho = \exp(-\beta \mathsf H)$. One can regard this formula as arising from the quantum mechanical entanglement with a heat bath which produces mixed states. In other words, FTTN regards it as thermal average loss computed from the entanglement with the environment instead of the direct summation. As for this classification task, the MPS tensor can be treated as the heat bath with a certain Fermi energy, and the individual local feature map entangled with the MPS tensor to produce mixed states. We idealized the statistical mechanism, the physical system has a specific history and environment. We equilibrate the system with weak coupling to a heat bath (MPS tensor), then moving coupling between the heat bath and the system. 

Here a typical set of states $\{\left |\boldsymbol \psi (\boldsymbol i) \right \rangle \}$ with probabilities $P(i)$ need to be chosen, it satisfies the fundamental proposition: 
\begin{equation}
\sum_i P(i) \left | \boldsymbol \psi (\boldsymbol i) \right \rangle \left \langle \boldsymbol \psi (\boldsymbol i) \right | = \exp (-\beta \mathsf H)
\label{eqn.thermal}
\end{equation}
From this equation, the expectation value of any Hermitian operator $\mathsf H$ can be determined from average of  $\left \langle \boldsymbol \psi (\boldsymbol i) \right | \mathsf H \left | \boldsymbol \psi (\boldsymbol i) \right \rangle$, where each $\left | \boldsymbol \psi (\boldsymbol i) \right \rangle$ is chosen randomly according to $P(\boldsymbol i)$. Let $\left |\boldsymbol  i \right \rangle$ to be a complete orthonormal basis, then one specific states satisfies Equation~\ref{eqn.thermal} can be:
\begin{equation}
\begin{aligned}
\left | \boldsymbol \psi (\mathsf i) \right \rangle &= P(\boldsymbol i) ^ {-1/2} \exp (-\beta \mathsf H/2) \left | \boldsymbol i \right \rangle\\
P(i) &= \left \langle \boldsymbol i \right | \exp (-\beta \mathsf H) \left | \boldsymbol i \right \rangle = \text{trace} (\mathsf \rho \left | \boldsymbol i \right \rangle \left \langle \boldsymbol i \right | )\\
\end{aligned}
\end{equation}
Notice these results do not require the states $\{ \left |\boldsymbol  i\right \rangle \}$ to be the orthonormal basis. From the physical aspect, the final result should be divided by a partition function $Z = Tr(\rho ) = \sum_i P(i)$, but here it is omitted since the final result is determined by the largest one.

The insertion of our proposed temperature layer is in analog to this METTS algorithm. The contraction of the temperature layer and the feature map gives a thermal perturbation to inputs. Here we give an intuitive explaination, if a picture contains a perfectly straight line, both humans and computers can easily distinguish it. However, if the line is slightly bent, humans can still treat it as a straight line. But for computers, by accurate calculation, the line is treated as a curve rather than a line. For thermal fluctuation, the bent line is treated as the superposition state of line and curve, which makes the network structure treat it as line by ignoring the small perturbation. In other words, the temperature-like parameter $\beta$ is an analog to a threshold. If the difference between input and feature is smaller than the threshold, they are treated as the same; however, for input and features with difference larger than the threshold, they are separated.

Mathematically, for a small $\beta \rightarrow 0$, this temperature layer approaches the identity matrix, corresponding to the MPS machine learning structure without the temperature layer. However, for a large $\beta$, due to the exponential decay, the variation between features is exponentially flattened, resulting in an entirely black image. A proper $\beta$ emphasizes the main feature and by choosing a proper temperature, the image contrast is adjusted to produce easy-to-train images.


\section{Experiment}
\label{sec.result}
In this section, we implement FTTN using TensorNetwork \cite{roberts2019tensornetwork} with TensorFlow \cite{tensorflow} backend. Here we focus on the result on Fashion-MNIST due to its large variance, while FTTN also demonstrates improvement on other small-variance datasets like MNIST.

For TN without thermal fluctuation, it is observed that training with automatic gradients is nearly independent of the used bond dimension \cite{efthymiou2019tensornetwork}, while the prior bond dimension matters when using the DMRG-like sweeping algorithm \cite{stoudenmire2016supervised}. One possible reason is that the latter one changes the bond dimension during the optimization process, while the previous one does not. Therefore, we adapt the automatic gradient descent method rather than the DMRG-like sweeping algorithm, since we want to focus on the temperature-like parameter $\beta$ instead of other parameters like bond dimension.

We train on the total fashion-MNIST training set consisting of 60,000 images of size $28\times 28$. The test accuracy is calculated on the whole test dataset consisting of 10,000 images. To illustrate the effect of the temperature layer, we use the same setup as \cite{efthymiou2019tensornetwork}, where the typical setting that performs well is to use Adam optimizer \cite{kingma2014adam} with a learning rate of $10^{-4}$ and the batch size of 50.   We adapt the multi-class cross-entropy as loss function, which is defined in Equation~\ref{eqn.LossFunc}. The feature map is $\psi(x)=[1-p \quad p]^T$. We also evaluate our method to MNIST dataset, which has relatively small variation.

\begin{figure}[htb] 
    \center{\includegraphics[width=8cm]  {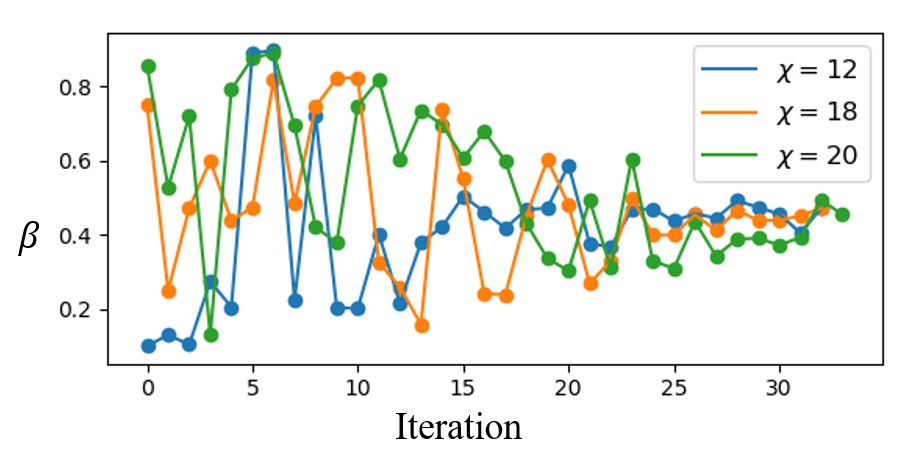}} 
    \caption{\label{fig.ConvGraph} The optimization process of the temperature-like parameter $\beta$.} 
\end{figure}

Ideally, the performance of FTTN achieves best when the given temperature is the same as that of the dataset. Here we adopt a simulated annealing algorithm to do the automatically optimization since they share a similar physical process. FTTN is given a randomly selected temperature. Then for each step, the parameter $\beta$ close to the current one is selected and its corresponding accuracy is measured, then this algorithm decides whether move to it or not depending on the temperature and the fact whether the new solution is better or worse than the previous one. The optimization process of this temperature-like parameter $\beta$ is shown in Figure~\ref{fig.ConvGraph}. It can be seen that the temperature-like hyper-parameter $\beta$ converges to $0.4$ for fashion-MNIST, independent of the bond dimension. 
Such parameter might be the intrinsic property of a dataset, and obtaining such property can be useful in training machine-learning-base tensor network frameworks.

\begin{figure}[htb] 
    \center{\includegraphics[width=8cm]  {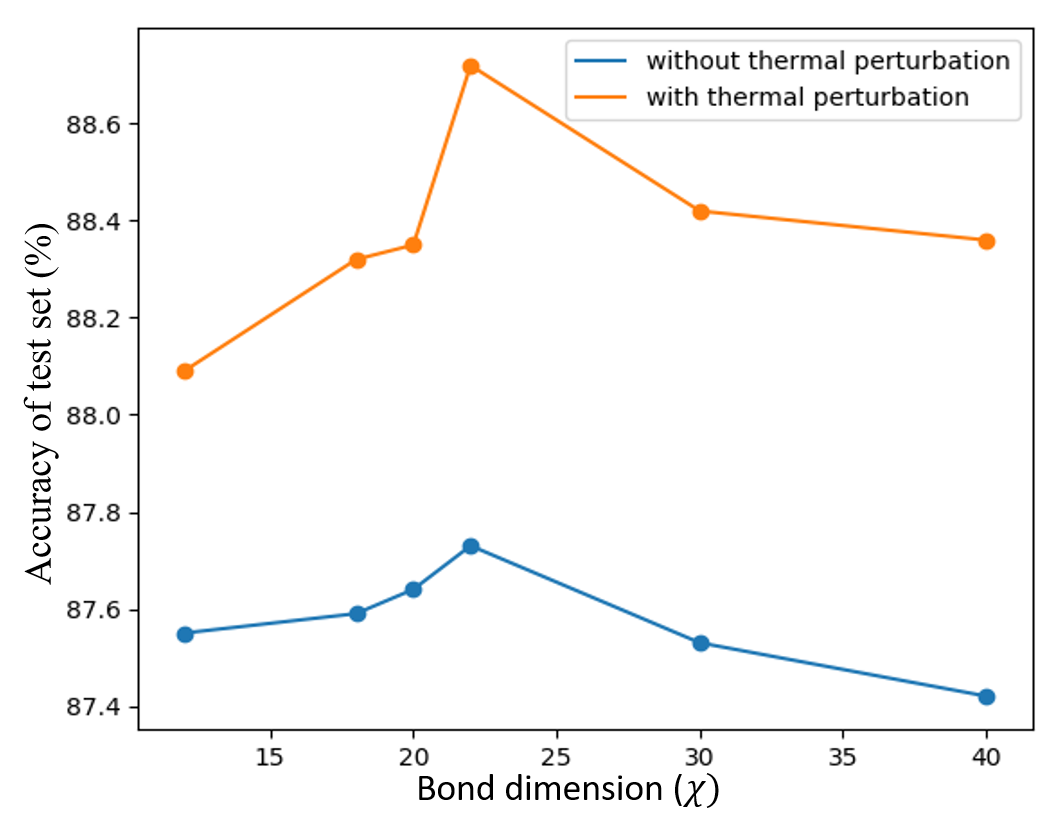}} 
    \caption{\label{fig.KaiValAcc} The $\chi$ dependent bond accuracy.Orange and blue lines correspond to the testing accuracy for Fashion-MNIST using tensor network with/without thermal perturbation respectively.} 
\end{figure}

Experiments observe that FTTN has following advantages:\\
\textbf{Test Accuracy}
As shown in Figure~\ref{fig.KaiValAcc}, we analyze the effect of thermal perturbation with different bond dimensions $\chi$ as shown in Figure~\ref{fig.KaiValAcc}. Without thermal perturbation, the test accuracy increase with bond dimension $\chi$ at first, but it goes through a decrease after a critical number. The introduced thermal perturbation does not change this tendency, but it can result in an overall improvement in test accuracy. Similar to what observed without thermal fluctuation, FTTN has tiny dependence on the bond dimension. We have seen that the thermal perturbation gives rise to the training accuracy and the convergence speed. The endorsement of the temperature layer gives an accuracy rising from $87.73\%$ to $88.72\%$, a one percentage increment 
for the database Fashion-MNIST. This means FTTN is suitable for this database since it has a large variation. However, even for the small variation database MNIST, FTTN also gives an accuracy increment from $98.31\%$ to $98.43\%$. 

\begin{figure}[htb] 
    \center{\includegraphics[width=8cm]  {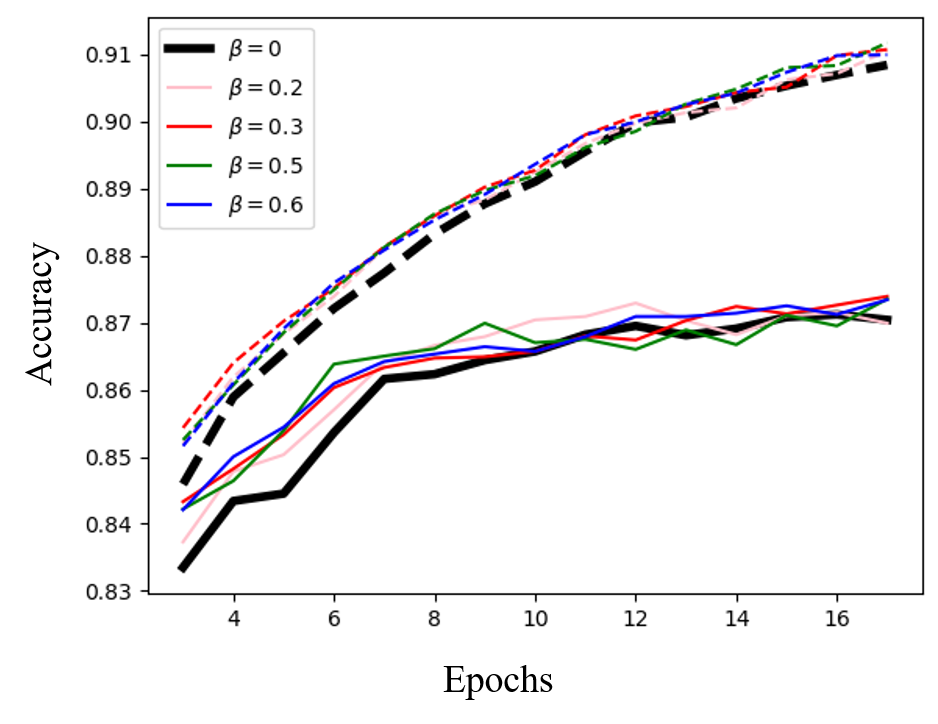}} 
    \caption{\label{fig.conspeed} The first 3-18 epoch of FTTN with different $\beta$. Here we use $\chi =12$ as an example, while similar situation happens for other $\chi $ value. The solid and dashed line represent the accuracy for test and training set respectively.} 
\end{figure}

\textbf{Speed of Convergence} We observed that the temperature layer can improve the speed of convergence. The speed of convergence is significantly improved in the first several epochs as shown in Figure~\ref{fig.conspeed}. Here the bond dimension $\chi$ is set as 12. We can see that the proper $\beta$ gives a larger than 2\% increment on the testing accuracy during the first several epoches. From our experiment, even though the improvement will decreases with increasing epoches, the improvement still possesses a considerable quantity.

\begin{figure}[htb] 
    \center{\includegraphics[width=7.5cm]  {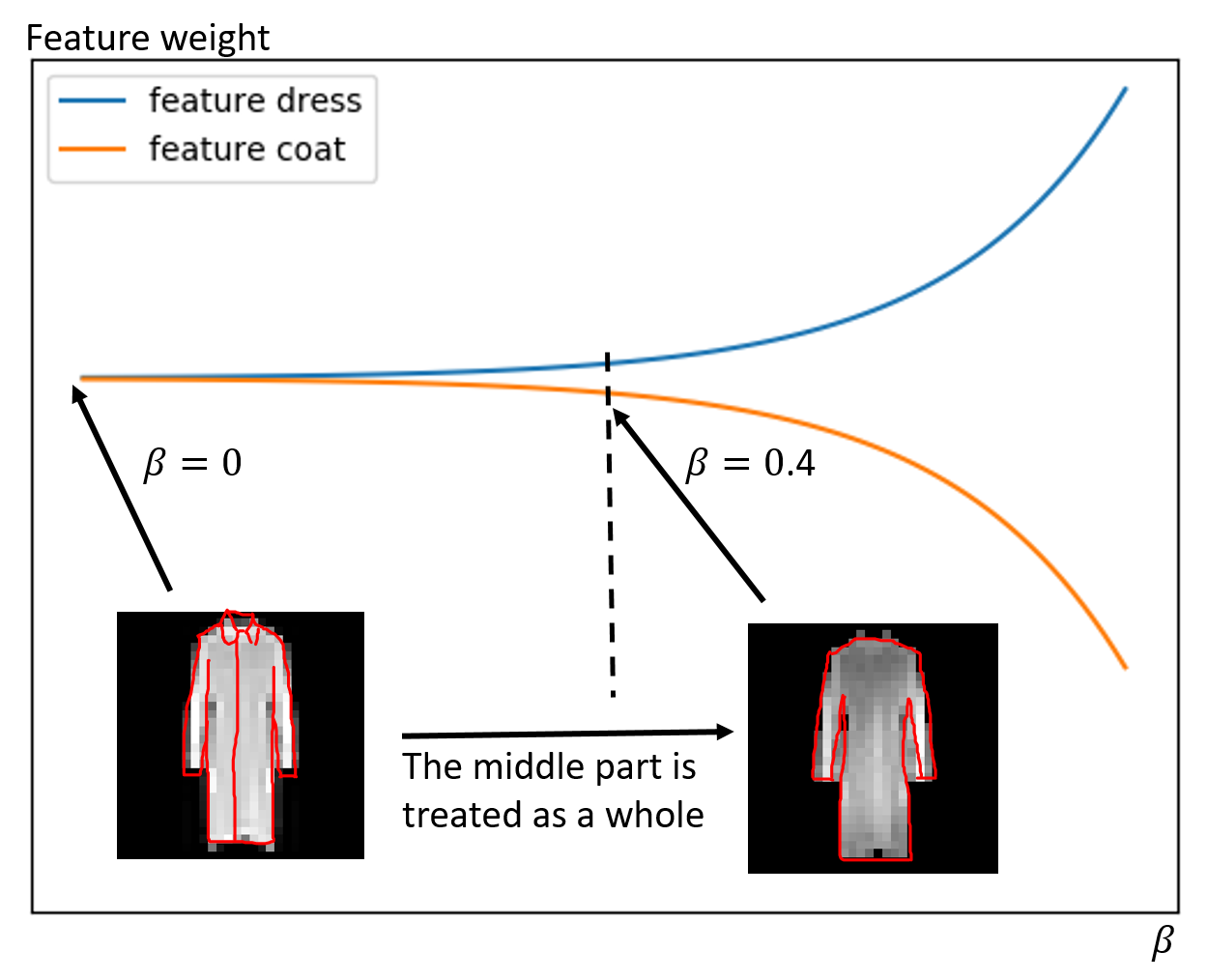}} 
    \caption{\label{fig.Fluctuation} The schematic diagram of the thermal fluctuation effect. The left figure is adapt from fashion-MNIST, while the right one is this image with thermal fluctuation. The perturbation separates similar features of  ``coat'' and ``dress'' to make a correct classification. }
\end{figure}

The thermal perturbation effect is illustrated in Figure~\ref{fig.Fluctuation}. The photo shown in Figure~\ref{fig.Fluctuation} is inaccurately classified as ``coat'' without perturbation, while the thermal fluctuation gives it a correct label ``dress''. We tried the traditional CNN structure using two convolutional layers and two pooling layers \cite{Fashion-MNIST}, despite its higher accuracy, this image can not be accurately classified. One of the differences is that the middle section is separated for ``coat'' while for ``dress'', the entire middle part is treated as a whole. Initially at $\beta = 0$, the feature of dress and coat degenerated since they share similar features, here a line in the middle is added to emphasize their differences. However, these two features are separated with the increasing $\beta$ value, which is the effect of the temperature layer. A proper $\beta$ gives enough separation to similar features. 


\section{Conclusions}
In this work, we introduce a thermal perturbation into the tensor network structure by insertion of a temperature layer and then we analyze the effect of this thermal perturbation. The constructed structure is in analogue to that from METTS algorithm \cite{stoudenmire2010minimally}. Such modification leads to around one percentage increment of accuracy in Fashion-MNIST dataset. Meanwhile, the convergence speed is accelerated, which is significant in the first several epoches. We hope that this technique we proposed here will be taken up by the wider machine learning community.

While using an one-dimension MPS ansatz for the classification works well even for two-dimensional data as shown in this work, there still exist some other feasible tensor network structures, like projected entangled pair states (PEPS) \cite{verstraete2004renormalization},  multiscale entanglement renormalization ansatz (MERA) \cite{vidal2007entanglement}, comb tensor networks \cite{chepiga2019comb}. Our method of introducing thermal fluctuation may also be applied to these frameworks and gives an enhancement. There is also much room to improve the temperature-endorsed optimization algorithm by incorporating with other frameworks like convolutional neural network and graph neural network. Besides the thermal fluctuation, symmetry is excellent in improving tensor network \cite{vanhecke2019symmetric}. The recurrent neural network has been proven as an analogy to wave physics \cite{hughes2019wave}, and we believe the physics-endorsed frameworks will greatly contribute to the machine learning community.


\bibliographystyle{named}
\bibliography{article}

\end{document}